\documentclass{IOS-Book-Article}

\usepackage{times}
\normalfont
\usepackage[T1]{fontenc}
\usepackage{url,graphicx,latexsym,xspace,amssymb,amsmath}

\newcommand{\ie}{\textit{i.e.}}
\newcommand{\eg}{\textit{e.g.}}

\newcommand{\Datalog}{\textsc{Datalog}}
\newcommand{\negDatalog}{\textsc{Datalog}$^{\neg}$}

\newcommand{\disjDatalog}{\textsc{Datalog}$^{\vee}$}
\newcommand{\disjnegDatalog}{\textsc{Datalog}$^{\neg\vee}$}
\newcommand{\Carin}{\textsc{Carin}}
\newcommand{\ALlog}{$\mathcal{AL}$-\textsc{log}}
\newcommand{\DL}{$\mathcal{DL}$}
\newcommand{\DLlog}{\DL+\textsc{log}}
\newcommand{\SHIQlog}{$\mathcal{SHIQ}$+\textsc{log}}
\newcommand{\negDLlog}{\DL+\textsc{log}$^{\neg}$}
\newcommand{\disjDLlog}{\DL+\textsc{log}$^{\vee}$}
\newcommand{\disjnegDLlog}{\DL+\textsc{log}$^{\neg\vee}$}
\newcommand{\Foil}{\textsc{Foil}}
\newcommand{\ORFoil}{\textsc{OR}-\Foil}
\newcommand{\DLFoil}{\DL-\Foil}

\newtheorem{definition}{Definition} 
\newtheorem{example}{Example} 

\begin{document}

\begin{frontmatter}                           

\title{Learning Onto-Relational Rules with Inductive Logic Programming %
}
\runningtitle{Learning Onto-Relational Rules with ILP}

\author[A]{\fnms{Francesca A.} \snm{Lisi}%
\thanks{Corresponding Author: Francesca A. Lisi, Dipartimento di Informatica, Universit\`{a} degli Studi di Bari ``Aldo Moro", Italy; E-mail: lisi@di.uniba.it}},

\runningauthor{Lisi}
\address[A]{Dipartimento di Informatica, Universit\`{a} degli Studi di Bari ``Aldo Moro", Italy}

\begin{abstract}
Rules complement and extend ontologies on the Semantic Web. We refer to these rules as onto-relational since they combine DL-based ontology languages and Knowledge Representation formalisms supporting the relational data model within the tradition of Logic Programming and Deductive Databases. Rule authoring is a very demanding Knowledge Engineering task which can be automated though partially by applying Machine Learning algorithms. In this chapter we show how Inductive Logic Programming (ILP), born at the intersection of Machine Learning and Logic Programming and considered as a major approach to Relational Learning, can be adapted to Onto-Relational Learning. For the sake of illustration, we provide details of a specific Onto-Relational Learning solution to the problem of learning rule-based definitions of DL concepts and roles with ILP. 
\end{abstract}


\end{frontmatter}

\thispagestyle{empty}
\pagestyle{empty}

\section*{Introduction}

\emph{Rules} are widely used in Knowledge Engineering (KE) and Knowledge Representation (KR) as a powerful way of modeling knowledge. In the broadest sense, a rule could be any statement which says that a certain conclusion must be valid whenever a certain premise is satisfied, \ie\ any statement that could be read as a sentence of the form ``if .. then ..". Rules have been successfully applied in the fields of Logic Programming (LP) and Deductive Databases \cite{Ceri90}. 
Rules play also a role in the \emph{Semantic Web} architecture. Interest in this area has grown rapidly over recent years as testified by the Rules Interchange Format (RIF)\footnote{\texttt{http://www.w3.org/2005/rules/wiki/RIF\_Working\_Group}} activity at W3C. Rules from the RIF perspective would allow the integration, transformation and derivation of data from numerous sources in a distributed, scalable, and transparent manner. Because of the great variety in rule languages and rule engine technologies, RIF consists of a core language\footnote{\texttt{http://www.w3.org/TR/rif-core/}} to be used along with a set of standard and non-standard extensions. These extensions need not all be combinable into a single unified language. As for the expressive power, two directions are followed: monotonic extensions towards full First Order Logic (FOL) and non-monotonic (NM) extensions based on the LP tradition. The debate around a RIF has taken a long time also due to the controversial issue of having rules on top or aside ontologies \cite{HorrocksADKGW03}. 
There is a consensus now on the fact that rules complement and extend ontologies. Indeed, rules can be used in combination with ontologies, or as a means to specify ontologies. They are also frequently applied over ontologies, to draw inferences, express constraints, specify policies, react to events, discover new knowledge, transform data, etc. 
In particular, RIF rules can refer to RDF and OWL facts. 
Since the design of OWL has been based on the $\mathcal{SH}$ family of very expressive \emph{Description Logics} (DLs) (see Chapter \ref{OL-book-chapter-description-logics} for an introduction)
, the NM dialects of RIF will most likely be inspired by those hybrid KR systems that integrate DLs and LP. Such rule formalisms are of interest to this chapter. We shall refer to them as \emph{onto-relational rule languages} from now on. 
Apart from the specific ontology language, the integration of ontologies and rules is already present in existing knowledge bases (KBs). Notably the Cyc\footnote{\texttt{http://cyc.com/cyc/technology/whatiscyc\_dir/}} KB consists of terms (which constitute the vocabulary, \ie\ the ontology) and assertions which relate those terms and include both simple ground assertions and rules \cite{LenatGPPS90}. 

The acquisition of rules for very large KBs like Cyc is a very demanding KE activity. Indeed, according to an estimate from the Cyc project, human experts produce rules at the rate of approximately three per hour but can evaluate an average of twenty rules per hour. Also, for untrained knowledge engineers, while rule authoring may be very difficult, rule reviewing is feasible (although still difficult). A partial automation of the rule authoring task, \eg by applying \emph{Machine Learning} (ML) algorithms (see Chapter \ref{OL-book-chapter-machine-learning} for an introduction), can be of help even though the automatically produced rules are not guaranteed to be correct. In fact, of those rules, some will turn out to be correct, and some will be found to need editing to be assertible. Yet, as mentioned above, rule reviewing is less critical than rule authoring. 
In order to partially automate the authoring of onto-relational rules, the bunch of ML techniques collectively known under the name of \emph{Inductive Logic Programming} (ILP) \cite{Nienhuys97} seems particularly promising for the following reasons. ILP was born at the intersection of ML and LP \cite{Muggleton91a}, and is widely recognized as a major approach to \emph{Relational Learning} \cite{DeRaedt08}. 
Apart from the KR framework of LP, the distinguishing feature of ILP, also with respect to other ML forms, is the use of prior domain knowledge in the form of a logical theory during the induction process. In this chapter we take a critical look at ILP proposals for learning relational rules while having an ontology as the background theory. These proposals try to overcome the difficulties of accommodating ontologies in Relational Learning. The work of \cite{CabralKMWS05} on using semantic meta-knowledge from Cyc as inductive bias in an ILP system is another attempt at solving this problem though more empirically. 
Conversely, we promote an extension of Relational Learning, called \emph{Onto-Relational Learning} (ORL), which accounts for ontologies in a clear, elegant and well-founded manner by resorting to onto-relational rule languages. In this chapter, for the sake of illustration, we provide details of a specific ORL solution to the problem of learning rule-based definitions of DL concepts and roles with ILP. 

The chapter is organized as follows. Section \ref{sect:pre} is devoted to preliminaries on LP and its applications to databases and ontologies as well as on ILP. Section \ref{sect:ilp4sw-today} provides a state-of-the-art survey of ILP proposals for learning onto-relational rules. Section \ref{sect:learning-fwk} describes in depth the most powerful of these proposals. Section \ref{sect:concl} concludes the chapter with final remarks and outlines directions of future work. 

\section{Preliminaries}\label{sect:pre}

\subsection{Logic Programming and databases}\label{sect:lp-db}

Logic Programming (LP) is rooted into a fragment of Clausal Logics (CLs) known as Horn Clausal Logic (HCL) \cite{Lloyd87}. The basic element in CLs is the \emph{atom} of the form $p(t_{i}, \ldots,
t_{k_{i}})$ such that each $p$ is a predicate symbol and each
$t_{j}$ is a term. A \emph{term} is either a constant or a variable or a more complex term obtained by applying a functor to simpler term. Constant, variable, functor and predicate symbols belong to mutually disjoint alphabets.
A \emph{literal} is an atom either negated or not. A \emph{clause}
is a universally quantified disjunction of literals. 
Usually the universal quantifiers are omitted to simplify
notation. Alternative notations are a clause as set of literals
and a clause as an implication. A \emph{program} is a set of clauses. HCL admits only so-called definite clauses.
A \emph{definite clause} is an implication of the form
\begin{center}
$\alpha_{0} \leftarrow \alpha_{1},\ldots, \alpha_{m}$
\end{center}
where $m \geq 0$ and $\alpha_{i}$ are atoms, \ie\ a clause with
exactly one positive literal.
The right-hand side $\alpha_{0}$ and the left-hand side
$\alpha_{1},\ldots, \alpha_{m}$ of the implication are called
\emph{head} and \emph{body} of the clause, respectively. Note that
the body is intended to be an
existentially quantified conjunctive formula $\exists \alpha_{1}
\wedge \ldots \wedge \alpha_{m}$. Furthermore definite
clauses with $m > 0$ and $m = 0$ are called \emph{rules} and
\emph{facts} respectively. The model-theoretic semantics of HCL is based on the notion of \emph{Herbrand interpretation}, \ie\ an interpretation in which all all constants and function symbols are assigned very simple meanings. This allows the symbols in a set of clauses to be interpreted in a purely syntactic way, separated from any real instantiation. The corresponding proof-theoretic semantics is based on the \emph{Closed World Assumption} (CWA), \ie\ the presumption that what is not currently known to be true, is false. Deductive reasoning with HCL is formalized in its proof theory. In clausal logic \emph{resolution} comprises a single inference rule which, from any two clauses having an appropriate form, derives a new clause as their consequence. Resolution is sound: every resolvent is implied by its parents. It is also refutation complete: the empty clause is derivable by resolution from any set $S$ of Horn clauses if $S$ is unsatisfiable. \emph{Negation As Failure} (NAF) is related to the CWA, as it amounts to believing false every predicate that cannot be proved to be true. Clauses with NAF literals in the body are called \emph{normal clauses}. The concept of a \emph{stable model}, or \emph{answer set}, is used to define a declarative semantics for normal logic programs \cite{GelfondL91}. According to this semantics, a logic program may have several alternative models (but possibly none), each corresponding to a possible view of the reality. Also based on the stable model (answer set) semantics, \emph{Answer Set Programming} (ASP) is an alternative LP paradigm oriented towards difficult search problems \cite{MarekT99}. 

Definite clauses played a prominent role in the rise of deductive databases \cite{Ceri90}. More precisely, functor-free non-recursive definite clauses are at the basis of the language Datalog for deductive databases \cite{CeriGT89}. Generally, it is denoted by \negDatalog\ where $\neg$ is treated as NAF. The restriction of Datalog to only positive rules (\ie, rules without NAF literals) is denoted by \Datalog. Based on the distinction between extensional and intensional
predicates, a \Datalog\ program $\Pi$ can be divided into two
parts. The \emph{extensional
part}, denoted as $EDB(\Pi)$, is the set of facts of $\Pi$ involving the
extensional predicates, whereas the \emph{intensional part}
$IDB(\Pi)$ is the set of all other clauses of $\Pi$. The main reasoning task in \Datalog\ is \emph{query answering}. A \emph{query} $Q$ to a \Datalog\ program $\Pi$ is a
\Datalog\ clause of the form
\begin{center}
$\leftarrow \alpha_{1},\ldots, \alpha_{m}$
\end{center} where $m > 0$, and $\alpha_{i}$ is a \Datalog\ atom. An \emph{answer} to a query $Q$ is a substitution $\theta$ for the
variables of $Q$. An answer is correct with respect to the
\Datalog\ program $\Pi$ if $\Pi \models Q \theta$. The
\emph{answer set} to a query $Q$ is the set of answers to $Q$ that
are correct w.r.t. $\Pi$ and such that $Q \theta$ is ground. In
other words the answer set to a query $Q$ is the set of all ground
instances of $Q$ which are logical consequences of $\Pi$. Answers are computed by refutation.

Disjunctive Datalog (denoted as \disjDatalog) is a variant of \Datalog\ where disjunctions may appear in the rule heads \cite{EiterGM97}. Therefore \disjDatalog\ can not be considered as a fragment of HCL. Advanced versions (\disjnegDatalog) also allow for negation in the bodies, which can be handled according to a semantics for negation in CLs. 
Defining the semantics of a \disjnegDatalog\ program is complicated by the presence of disjunction in the rules' heads because it makes the underlying disjunctive logic programming inherently nonmonotonic, \ie\ new information can invalidate previous conclusions. Among the many alternatives, one widely accepted semantics for \disjnegDatalog\ is the extension of the stable model semantics to the disjunctive case. 

\subsection{Logic Programming and ontologies}\label{sect:lp-sw}

The integration of LP and ontologies follows the tradition of KR research on so-called \emph{hybrid systems}, \ie\ those systems which are constituted by two or more subsystems dealing with distinct portions of a single KB by performing specific reasoning procedures \cite{FrischC91}. 
The motivation for investigating and developing such systems is to improve on two basic features of KR formalisms, namely \emph{representational adequacy} and \emph{deductive power}, by preserving the other crucial feature, \ie\ \emph{decidability}. Indeed DLs and CLs are FOL fragments incomparable as for the expressiveness \cite{Borgida96} and the semantics \cite{Rosati05} but combinable at different degrees of integration: Tight, loose, full.  

The semantic integration is \emph{tight} when a model of the hybrid KB is defined as the union of two models, one for the DL part
and one for the CL part, which share the same domain. In particular, combining DLs with CLs in a tight manner can easily lead to undecidability if the interaction scheme between the DL and the CL part of a hybrid KB does not solve the semantic mismatch between DLs and CLs \cite{Rosati05-ppswr}. This requirement is known as \emph{DL-safety} \cite{MotikSS05}. With respect to this property, the hybrid KR system \Carin\ \cite{LevyR98} is \emph{unsafe} because the interaction scheme is left unrestricted. Conversely, \ALlog\ \cite{Donini98} guarantees a \emph{safe} interaction scheme by means of syntactic restrictions. Finally, \disjnegDLlog\ \cite{Rosati06}\footnote{We prefer \disjnegDLlog\ to the original name \DLlog\ in order to emphasize the NM features of the language.} is \emph{weakly DL-safe} because it relaxes the condition of DL-safety. The distinguishing features of these three KR frameworks are summarized in Table \ref{tab:kr-comp} and further discussed in Section \ref{sect:carin}, \ref{sect:al-log}, and \ref{sect:dl+log} respectively. 
\begin{table*}[t]
\caption{Three KR frameworks suitable for representing onto-relational rules.}\label{tab:kr-comp}
\begin{center}
\tiny
\begin{tabular}{r@{\hspace{.2cm}}|p{3.5cm}|p{3.0cm}|p{3.0cm}}
 & \textbf{\textsc{Carin}} \cite{LevyR98} & \textbf{\ALlog}\cite{Donini98} & \textbf{\disjnegDLlog}\cite{Rosati06}\\
\hline
\noalign{\smallskip}
DL language & any DL & $\mathcal{ALC}$ & any DL \\
CL language & Horn clauses & \Datalog\ clauses & \disjnegDatalog\ clauses\\
\hline
\noalign{\smallskip}
integration & tight DL-unsafe & tight DL-safe & tight weakly DL-safe\\
rule head literals & DL/Horn literals & \Datalog\ literal & DL/\Datalog\ literals\\
rule body literals & DL/Horn literals & $\mathcal{ALC}$/\Datalog\ literals (no roles)& DL/\negDatalog\ literals\\
\hline
\noalign{\smallskip}
semantics & Herbrand models+DL models & idem & stable models+DL models\\
reasoning & SLD-resolution+tableau calculus & idem & stable model computation + Boolean CQ/UCQ containment \\
decidability & only for some instantiations & yes & for all instantiations with DLs for which the Boolean CQ/UCQ containment is decidable\\
\hline
\noalign{\smallskip}
implementation & yes, \eg \cite{DBLP:journals/ijcis/GoasdoueLR00} & yes, \eg \cite{RuckhausKP06} & unknown\\
\hline
\end{tabular}
\end{center}
\end{table*}

The semantic integration is \emph{loose} when the DL part and the CL part are separate components connected through a minimal interface for exchanging
knowledge. An example of one such kind of coupling is the integration scheme for ASP and DLs illustrated in \cite{EiterILST08}. It derives from the previous work of the same authors on the extension of ASP with higher-order reasoning and external evaluations \cite{DBLP:conf/ijcai/EiterIST05} which has been implemented into the system DLVHEX\footnote{\texttt{http://www.kr.tuwien.ac.at/research/systems/dlvhex/}}. 

The semantic integration is \emph{full} when there is no separation between vocabularies of the two parts of the hybrid KB. One such kind of coupling is achieved by means of the logic of Minimal Knowledge and Negation as Failure in \cite{DBLP:journals/jacm/MotikR10}. 

A complete picture of the computational properties of systems combining DL ontologies and
\Datalog\ rules can be found in \cite{Rosati08}. An updated survey of the literature on hybrid DL-CL systems \cite{DBLP:series/lncs/DrabentEIKLM09} is suggested for further reading. 

\subsubsection{\Carin}\label{sect:carin}

A comprehensive study of the effects of combining DLs and CLs (more precisely, Horn rules) can be found in \cite{LevyR98}. Special attention is devoted to the DL $\mathcal{ALCNR}$. The results of the study can be summarized as follows: (i) answering conjunctive queries over $\mathcal{ALCNR}$ TBoxes is decidable, (ii) query answering in $\mathcal{ALCNR}$ extended with non-recursive \Datalog\ rules, where both concepts and roles can occur in rule bodies, is also decidable, as it can be reduced to answering a \emph{union of conjunctive queries} (UCQ)\footnote{A UCQ over a predicate alphabet $P$ is a FOL sentence of the form $\exists \vec{X}.conj_1(\vec{X})\vee \ldots \vee conj_n(\vec{X})$, where $\vec{X}$ is a tuple
of variable symbols and each $conj_i(\vec{X})$ is a set of atoms whose predicates are in $P$ and whose arguments are either
constants or variables from $\vec{X}$. A CQ is a UCQ with $n = 1$.}, (iii) if rules are recursive, query answering becomes undecidable, (iv) decidability can be regained by disallowing certain combinations of constructors in the logic, and (v) decidability can be regained by requiring rules to be \emph{role-safe}, where at least one variable from each role literal must occur in some non-DL-atom. The integration framework proposed in \cite{LevyR98} and known as \textsc{Carin} is therefore DL-unsafe. Reasoning in \textsc{Carin} is based on \emph{constrained SLD-resolution}, \ie\ an extension of SLD-resolution with a tableau calculus for DLs to deal with DL literals in the rules. Constrained SLD-refutation is a complete and sound method for answering \emph{ground} queries. 

\subsubsection{\ALlog}\label{sect:al-log}

\ALlog\ is a hybrid KR system that integrates safely the DL $\mathcal{ALC}$ and \Datalog\ \cite{Donini98}. In particular, variables occurring in the body of rules may be constrained with $\mathcal{ALC}$ concept assertions to be used as 'typing constraints'. This makes rules applicable only to explicitly named objects. As in \Carin, query answering is decided using the constrained SLD-resolution which however in \ALlog\ is decidable and runs in single non-deterministic exponential time. 

\subsubsection{\disjnegDLlog}\label{sect:dl+log}

The hybrid KR framework of \disjnegDLlog\ allows a \DL\ KB, \ie\ a KB expressed in any DL, to be extended with weakly DL-safe \disjnegDatalog\ rules \cite{Rosati06}. Weak DL-safeness allows to overcome the main representational limits of the DL-safe approaches, \eg\ the possibility of expressing UCQs, by keeping the integration scheme still decidable. For \disjnegDLlog\ two semantics have been defined: a FOL semantics and a NM semantics. In particular, the latter extends the stable model semantics of \disjnegDatalog. According to it, \DL-predicates are still interpreted under OWA, while \Datalog-predicates are interpreted under CWA. Notice that, under both semantics, entailment can be reduced to satisfiability and, analogously, that CQ answering can be reduced to satisfiability. The problem statement of satisfiability for finite \disjnegDLlog\ KBs relies on the problem known as the \emph{Boolean CQ/UCQ containment problem}\footnote{This problem was called \emph{existential entailment} in \cite{LevyR98}.} in \DL. It is shown that the decidability of reasoning in \disjnegDLlog, thus of ground query answering, depends on the decidability of the Boolean CQ/UCQ containment problem in \DL. Currently, $\mathcal{SHIQ}$ is one of the most expressive DLs for which this problem is decidable \cite{GlimmHLS08}.

\subsection{Inductive Logic Programming}\label{sect:ilp}

Inductive Logic Programming (ILP) was born at the intersection between LP and ML \cite{Muggleton91a}. From LP it has borrowed the KR framework, \ie\ HCL. From ML (more precisely, from Concept Learning) it has inherited the inferential mechanisms for induction, the most prominent of which is \emph{generalization}. However, a distinguishing feature of ILP with respect to other forms of Concept Learning is the use of prior knowledge of the domain of interest, called \emph{background knowledge} (BK). Therefore, induction with ILP generalizes from individual instances/observations in the presence of BK, finding valid hypotheses. \emph{Validity} depends on the underlying \emph{setting}. At present, there exist several formalizations of
induction in ILP that can be classified according to the following two orthogonal dimensions: the \emph{scope of induction} (discrimination vs
characterization) and the \emph{representation of observations} (ground definite
clauses vs ground unit clauses). \emph{Discriminant induction} aims at
inducing hypotheses with discriminant power as required in tasks like
classification. In classification, observations encompass both positive and
negative examples. \emph{Characteristic induction} is more suitable for finding regularities
in a data set. This corresponds to learning from positive
examples only. 
The second dimension affects the notion of \emph{coverage}, \ie\ the condition under which a hypothesis explains an observation. In \emph{learning from entailment}, hypotheses are clausal theories, observations are ground definite clauses, and a
hypothesis covers an observation if the hypothesis logically entails the observation. In \emph{learning
from interpretations}, hypotheses are clausal theories, observations are
Herbrand interpretations (ground unit clauses) and a hypothesis covers an observation if the observation is a model for the hypothesis. 

In Concept Learning, generalization is traditionally viewed as search through a partially ordered space of inductive hypotheses \cite{Mitchell82}. According to this vision, an inductive hypothesis in ILP is a clausal theory and the induction of a single clause requires (i) structuring, (ii) searching and (iii) bounding the space of clauses \cite{Nienhuys97}. 
First we focus on (i) by clarifying the notion of \emph{ordering} for clauses. An ordering allows for determining which one, between two clauses, is more general than the other. Since partial orders are considered, uncomparable pairs of clauses are admitted. 
Given the usefulness of BK, orders have been proposed that reckon
with it. Among them, \emph{generalized subsumption} \cite{Buntine88} is of major interest to this chapter: Given two definite clauses $C$ and $D$ standardized apart and a definite program $\mathcal{K}$, we say that $C \succeq_{\mathcal{K}} D$ iff there
exists a ground substitution $\theta$ for $C$ such that (i)
$head(C)\theta=head(D)\sigma$ and (ii) $\mathcal{K} \cup body(D)\sigma \models
body(C)\theta$ where $\sigma$ is a Skolem substitution
for $D$ with respect to $\{C\} \cup \mathcal{K}$. 
Generalized subsumption is also called \emph{semantic generality} in contrast to other orders which are purely syntactic. In the general case, it is undecidable. 
However, for \Datalog\ it is decidable and admits a least general generalization.
Once structured, the space of hypotheses can be searched (ii) by means of refinement operators. 
A \emph{refinement operator} is a function which computes a set of specializations or generalizations of a clause according to whether a top-down or a bottom-up search is performed. The two kinds of refinement operator have been therefore called \emph{downward} and \emph{upward}, respectively. The definition of refinement operators
presupposes the investigation of the properties of the various orderings and is usually coupled with the specification of a declarative bias for bounding the space of clauses (iii). \emph{Bias} concerns anything which
constrains the search for theories, e.g. a \emph{language bias} specifies syntactic constraints such as \emph{linkedness} and \emph{connectedness} on the clauses in the search space. A definite clause $C$ is linked if each literal $l_{i} \in C$ is linked. A literal $l_{i} \in C$ is linked if at least one of its terms is linked. A term $t$ in some literal $l_{i}
\in C$ is linked with linking-chain of length 0, if $t$ occurs in $head(C)$,
and with linking-chain of length $d+1$, if some other term in $l_{i}$ is
linked with linking-chain of length $d$. The link-depth of a term $t$ in $l_{i}$ is the length of the shortest linking-chain of $t$. A clause $C$ is connected if each variable occurring in $head(C)$ also occurs in $body(C)$.

\section{ILP for Onto-Relational Rule Learning: State of the Art}\label{sect:ilp4sw-today}

Hybrid KR systems combining DLs and CLs with a tight integration scheme have very recently attracted some attention in the ILP community: \cite{RouveirolV2000} chooses \textsc{Carin}-$\mathcal{ALN}$, \cite{Lisi08} resorts to \ALlog, and \cite{LisiE08-ilp} builds upon \SHIQlog. A comparative analysis of the three is reported in Table \ref{tab:ilp-comp}. They can be considered as attempts at accommodating ontologies in ILP. Indeed, they can deal with $\mathcal{ALN}$, $\mathcal{ALC}$, and $\mathcal{SHIQ}$ ontologies respectively. We remind the reader that $\mathcal{ALN}$ and $\mathcal{ALC}$ are incomparable DLs whereas DLs in the $\mathcal{SH}$ family enrich $\mathcal{ALC}$ with further constructors. 

Closely related to KR systems integrating DLs and CLs are the hybrid formalims arising from the study of
many-sorted logics, where a FOL language is combined with a sort language
which can be regarded as an elementary DL \cite{Frisch91}. In this respect the study
of a sorted downward refinement \cite{Frisch99} can be also considered as a contribution to
the problem of interest to this chapter. Finally, some work has been done on discovering frequent association patterns in the form of DL-safe rules \cite{JozefowskaLL10}. 

\subsection{Learning \Carin-$\mathcal{ALN}$ rules}

The framework proposed in \cite{RouveirolV2000} focuses on discriminant induction and adopts the ILP setting of learning from interpretations. Hypotheses are represented as \Carin-$\mathcal{ALN}$ non-recursive rules with a Horn literal in the head that plays the role of target concept. The coverage relation of hypotheses against examples adapts the usual one in learning from interpretations to the case of hybrid \Carin-$\mathcal{ALN}$ BK. The generality relation between two hypotheses is defined as an extension of generalized subsumption. Procedures for testing both the coverage relation and the generality relation are based on the existential entailment algorithm
of \Carin. Following \cite{RouveirolV2000}, Kietz studies the learnability of
\Carin-$\mathcal{ALN}$, thus providing a pre-processing method which enables
ILP systems to learn \Carin-$\mathcal{ALN}$ rules \cite{Kietz03}.

\subsection{Learning \ALlog\ rules}

In \cite{Lisi08}, hypotheses are represented as constrained \Datalog\ clauses that are linked, connected (or
range-restricted), and compliant with the bias of
Object Identity (OI)\footnote{The OI bias can be considered as an extension of the UNA from the semantic level to the syntactic one of \ALlog. It can be the starting point for the definition of
either an equational theory or a quasi-order for constrained \Datalog\
clauses.}. Unlike \cite{RouveirolV2000}, this framework is general, meaning that it is valid whatever the scope of induction is. 
The generality relation for one such hypothesis language is an adaptation of generalized subsumption, named $\mathcal{B}$-subsumption, to the \ALlog\ KR framework. It gives raise to a quasi-order and can be checked with a decidable procedure based on constrained SLD-resolution \cite{LisiM03-aiia}. Coverage relations for both ILP settings of learning from interpretations and learning from entailment have been defined on the basis of query answering in \ALlog\ \cite{LisiE04-ilp}. 
As opposed to \cite{RouveirolV2000}, the framework has been implemented in an ILP system \cite{LisiM04,DBLP:journals/ijswis/Lisi11}. More precisely, an instantiation of it for the case of \emph{characteristic induction from interpretations} has been considered. Indeed, the system supports a variant of a very popular data mining task - frequent pattern discovery - where rich prior conceptual knowledge is taken into account during the discovery process in order to find patterns at multiple levels of description granularity. The search through the space of patterns represented as unary conjunctive queries in \ALlog\ and organized according to $\mathcal{B}$-subsumption is performed by applying an ideal downward refinement operator \cite{LisiM03-ilp}.

\subsection{Learning \SHIQlog\ rules}

The ILP framework presented in \cite{LisiE08-ilp} represents hypotheses as \SHIQlog\ rules and organizes them according to a generality ordering inspired by generalized subsumption. The resulting hypothesis space can be searched by means of refinement operators either top-down or bottom-up. Analogously to \cite{Lisi08}, this framework encompasses both scopes of induction but, differently from \cite{Lisi08}, it assumes the ILP setting of learning from entailment only. Both the coverage relation and the generality relation boil down to query answering in \SHIQlog, thus can be reformulated as satisfiability problems. Compared to \cite{RouveirolV2000} and \cite{Lisi08}, this framework shows an added value which can be summarized as follows. First, it relies on a more expressive DL, \ie\ $\mathcal{SHIQ}$. Second, it allows for inducing definitions for new DL concepts, \ie\ rules with a $\mathcal{SHIQ}$ literal in the head. Third, it adopts a more flexible form of integration between the DL and the CL part, \ie\ the weakly-safe one. 

The work reported in \cite{LisiE09-ilp,DBLP:journals/tplp/Lisi10} generalizes the results of \cite{LisiE08-ilp} to any decidable instantiation of \disjnegDLlog. The following section illustrates how learning \negDLlog\ rules can support the evolution of ontologies.

\begin{table*}
\caption{Three ILP frameworks suitable for learning onto-relational rules.}\label{tab:ilp-comp}
\begin{center}
\tiny
\begin{tabular}{r@{\hspace{.2cm}}|l@{\hspace{.2cm}}|l@{\hspace{.2cm}}|l}
 & \textbf{Learning \Carin-$\mathcal{ALN}$ rules} \cite{RouveirolV2000} & \textbf{Learning \ALlog\ rules} \cite{Lisi08} & \textbf{Learning \SHIQlog\ rules} \cite{LisiE08-ilp}\\
\hline
\noalign{\smallskip}
prior knowledge  & \Carin-$\mathcal{ALN}$ KB & \ALlog\ KB & \SHIQlog\ KB \\
ontology language & $\mathcal{ALN}$ & $\mathcal{ALC}$ & $\mathcal{SHIQ}$\\
rule language & HCL & \Datalog\ & \Datalog\ \\
hypothesis language & \Carin-$\mathcal{ALN}$ non-recursive rules & \ALlog\ non-recursive rules & \SHIQlog\ non-recursive rules\\
target predicate & Horn predicate & \Datalog\ predicate & $\mathcal{SHIQ}$/\Datalog\ predicate\\
\hline
\noalign{\smallskip}
logical setting & interpretations & interpretations/entailment & entailment\\
scope of induction & prediction & prediction/description & prediction/description\\
\hline
\noalign{\smallskip}
generality order & extension of \cite{Buntine88} to \Carin-$\mathcal{ALN}$ & extension of \cite{Buntine88} to \ALlog\ & extension of \cite{Buntine88} to \SHIQlog \\
coverage test & \Carin\ query answering & \ALlog\ query answering & \disjnegDLlog\ query answering\\
ref. operators & n.a. & downward & downward/upward\\
\hline
\noalign{\smallskip}
implementation & unknown & yes, see \cite{DBLP:journals/ijswis/Lisi11} & no\\
application & no & yes, see \cite{LisiM04} & no\\
\hline
\end{tabular}
\end{center}
\end{table*}

\section{Learning Rule-based Definitions of \DL\ Concepts and Roles with ILP}\label{sect:learning-fwk}

%
In KE, Ontology Evolution is the timely adaptation of an ontology to changed
business requirements, to trends in ontology instances and patterns of usage of the ontology-based application, as well as the consistent
management/propagation of these changes to dependent elements \cite{StojanovicMMS02}. As opposed to Ontology Modification, Ontology Evolution must preserve the consistency of the ontology. According to \cite{NoyK04} one can distinguish between conceptual, specification and representation changes. 

In this section we consider the conceptual changes of a \DL\ ontology due to extensional knowledge (\ie, facts of the instance level of the ontology) previously unknown but classified which may become available. In particular, we consider the task of defining new concepts or roles which provide the intensional counterpart of such extensional knowledge and show how this task can be reformulated as an ORL problem \cite{LisiE08-swap}. 
For example, the new facts \texttt{LONER(Joe)}, \texttt{LONER(Mary)}, and \texttt{LONER(Paul)} concerning known individuals
may raise the need for having a definition of the concept \texttt{LONER} in the ontology. One such definition can be learned from these facts together with prior knowledge about \texttt{Joe}, \texttt{Mary} and \texttt{Paul}, \ie\ facts concerning them and already available in the ontology. A crucial requirement is that the definition must be expressed as a \DL\ formula or similar. In the following we provide the means for learning rule-based definitions of \DL\ concepts/roles in the KR framework of \negDLlog. 


\subsection{The learning problem}\label{sect:problem}

We assume that a \DL\ ontology $\Sigma = \left\langle \mathcal{T}, \mathcal{A} \right\rangle$ is integrated with a \negDatalog\ database $\Pi$ to form a \negDLlog\ KB $\mathcal{B}$. The problem of inducing rule-based definitions of \DL\ concepts/roles that do not occur in $\mathcal{B}$ can be formalized as follows. 
\begin{definition}\label{def:learning-problem}
Given:
\begin{itemize}
	\item a \negDLlog\ KB $\mathcal{B}$ (\emph{background theory}) 
	\item a \DL\ predicate name $p$ (\emph{target predicate})
	\item a set $\mathcal{E}=\mathcal{E}^{+} \cup \mathcal{E}^{-}$ of \DL\ assertions that are either true or false for $p$ (\emph{examples})
	\item a set $\mathcal{L}$ of \negDLlog\ definitions for $p$ (\emph{language of hypotheses})
\end{itemize}
the problem of building a rule-based definition of $p$ is to induce a set $\mathcal{H} \subset \mathcal{L}$ (\emph{hypothesis}) of \negDLlog\ rules from $\mathcal{E}$ and $\mathcal{B}$ such that:
\begin{description}
	\item[Completeness] $\forall e \in \mathcal{E}^{+}: \mathcal{H}$ covers $e$ w.r.t. $\mathcal{B}$
	\item[Consistency] $\forall e \in \mathcal{E}^{-}: \mathcal{H}$ does not cover $e$ w.r.t. $\mathcal{B}$.
\end{description}
\end{definition} 

The \emph{background theory} $\mathcal{B}$ in Definition \ref{def:learning-problem} can be split into an intensional part $\mathcal{K}$ (\ie, the TBox $\mathcal{T}$ plus $IDB(\Pi)$) and an extensional part $\mathcal{F}$ (\ie, the ABox $\mathcal{A}$ plus $EDB(\Pi)$). Also we denote by $P_\mathcal{C}(\mathcal{B})$, $P_\mathcal{R}(\mathcal{B})$, and $P_\textsc{D}(\mathcal{B})$ the sets of concept, role and \Datalog\ predicate names occurring in $\mathcal{B}$, respectively. Note that $p \not\in P_\mathcal{C}(\mathcal{B}) \cup P_\mathcal{R}(\mathcal{B})$.

\begin{example}\label{ex:shiq+log-kb}
Suppose we have a \negDLlog\ KB $\mathcal{B}$ (adapted from \cite{Rosati06}) built upon the alphabets $P_\mathcal{C}(\mathcal{B}) = \{\texttt{RICH/1}, \texttt{UNMARRIED/1}\}$, $P_\mathcal{R}(\mathcal{B}) = \{\texttt{WANTS-TO-MARRY/2}, \texttt{LOVES/2}\}$, and $P_\textsc{D}(\mathcal{B}) = \{\texttt{famous/1}, \texttt{scientist/1}, \texttt{meets/3}\}$ and consisting of the following intensional knowledge $\mathcal{K}$:
\begin{tabbing}
  MM \= MMMMMM \= MM \= \kill
$[A1]$ \texttt{RICH}$\sqcap$\texttt{UNMARRIED} $\sqsubseteq \exists$ \texttt{WANTS-TO-MARRY$^{-}$.$\top$}\\
$[A2]$ \texttt{WANTS-TO-MARRY} $\sqsubseteq$ \texttt{LOVES}\\
$[R1]$ \texttt{RICH(X)} $\leftarrow$ \texttt{famous(X)}, \texttt{$\neg$scientist(X)}\\
$[R2]$ \texttt{happy(X)} $\leftarrow$ \texttt{famous(X)}, \texttt{WANTS-TO-MARRY(Y,X)}
\end{tabbing}
and the following extensional knowledge $\mathcal{F}$:
\begin{tabbing}
  MM \= MMMMMM \= MM \= \kill
\> \texttt{UNMARRIED(Mary)}\\
\> \texttt{UNMARRIED(Joe)}\\
\> \texttt{famous(Mary)}\\
\> \texttt{famous(Paul)}\\
\> \texttt{famous(Joe)}\\
\> \texttt{scientist(Joe)}\\
\> \texttt{meets(Mary,Paul,Italy)}\\
\> \texttt{meets(Mary,Joe,Germany)}\\
\> \texttt{meets(Joe,Mary,Italy)}
\end{tabbing}
that concerns the individuals \texttt{Mary}, \texttt{Joe}, \texttt{Paul}, \texttt{Italy}, and \texttt{Germany}. 
\end{example}

The \emph{hypothesis language} $\mathcal{L}$ in Definition \ref{def:learning-problem} is given as a set of declarative bias constraints. It allows for the generation of \negDLlog\ rules starting from three disjoint alphabets $P_\mathcal{C}(\mathcal{L}) \subseteq P_\mathcal{C}(\mathcal{B})$, $P_\mathcal{R}(\mathcal{L}) \subseteq P_\mathcal{R}(\mathcal{B})$, and $P_\textsc{D}(\mathcal{L}) \subseteq P_\textsc{D}(\mathcal{B})$. Also we distinguish between $P_\textsc{D}^{+}(\mathcal{L})$ and $P_\textsc{D}^{-}(\mathcal{L})$ in order to specify which \Datalog\ predicates can occur in positive and negative literals, respectively. More precisely, we consider \negDLlog\ rules of the form
\begin{equation}\label{eq:hypothesis-syntax}
p(\vec{X}) \leftarrow r_1(\vec{Y_1}), \ldots, r_m(\vec{Y_m}), s_1(\vec{Z_1}), \ldots, s_k(\vec{Z_k}), \neg u_1(\vec{W_1}), \ldots, \neg u_q(\vec{W_q})
\end{equation}
where $m, k, q \geq 0$, $p(\vec{X})$ and each $r_j(\vec{Y_j})$, $s_l(\vec{Z_l})$, $u_t(\vec{W_t})$ is an atom with $r_j \in P_\textsc{D}^{+}(\mathcal{L})$, $s_l \in P_\mathcal{C}(\mathcal{L}) \cup P_\mathcal{R}(\mathcal{L})$, and $u_t \in P_\textsc{D}^{-}(\mathcal{L})$. The admissible rules must be compliant with the following restrictions:
\begin{description}
	\item[\Datalog-safeness] every variable occurring in (\ref{eq:hypothesis-syntax})
must appear in at least one of the atoms $r_1(\vec{Y_1}), \ldots, r_m(\vec{Y_m}), s_1(\vec{Z_1}), \ldots, s_k(\vec{Z_k})$;
	\item[weak \DL-safeness] every head variable of (\ref{eq:hypothesis-syntax}) must appear
in at least one of the atoms $r_1(\vec{Y_1}), \ldots, r_m(\vec{Y_m})$.
\end{description}
which also guarantee that the conditions of linkedness and connectedness, usually assumed in ILP, are satisfied.

\begin{example}\label{ex:hyp-lang2}
Suppose that the target predicate is the \DL\ concept \texttt{LONER}. If $\mathcal{L}^\texttt{LONER}$ is defined over $P_D^{+}(\mathcal{L}^\texttt{LONER}) \cup P_D^{-}(\mathcal{L}^\texttt{LONER}) \cup P_\mathcal{C}(\mathcal{L}^\texttt{LONER})= \{\texttt{famous/1}\} \cup \{\texttt{happy/1}\} \cup \{\texttt{RICH/1}, \texttt{UNMARRIED/1}\}$, then the following \negDLlog\ rules 
\begin{tabbing}
  MMMMMM \= MMMMMM \= MM \= \kill
	$h_1^\texttt{LONER}$ \> \texttt{LONER(X)} $\leftarrow$ \texttt{famous(X)}\\
	$h_2^\texttt{LONER}$ \> \texttt{LONER(X)} $\leftarrow$ \texttt{famous(X)}, \texttt{UNMARRIED(X)}\\
	$h_3^\texttt{LONER}$ \> \texttt{LONER(X)} $\leftarrow$ \texttt{famous(X)}, $\neg$\texttt{happy(X)}
\end{tabbing}  
belong to $\mathcal{L}^\texttt{LONER}$ and represent hypotheses of a definition for \texttt{LONER}. 
\end{example}
\begin{example}\label{ex:hyp-lang3}
Suppose now that the \DL\ role \texttt{LIKES} is the target predicate and the set $P_D^{+}(\mathcal{L}^\texttt{LIKES}) \cup P_\mathcal{C}(\mathcal{L}^\texttt{LIKES}) \cup P_\mathcal{R}(\mathcal{L}^\texttt{LIKES})=\{\texttt{happy/1}, \texttt{meets/3}\} \cup \{\texttt{RICH/1}\} \cup \{\texttt{LOVES/2}, \texttt{WANTS-TO-MARRY/2}\}$ provides the building blocks for the language $\mathcal{L}^\texttt{LIKES}$. The following \negDLlog\ rules 
\begin{tabbing}
  MMMMMM \= MMMMMM \= MM \= \kill
	$h_1^\texttt{LIKES}$ \> \texttt{LIKES(X,Y)} $\leftarrow$ \texttt{meets(X,Z,Y)}\\
  $h_2^\texttt{LIKES}$ \> \texttt{LIKES(X,Y)} $\leftarrow$ \texttt{meets(X,Z,Y)}, \texttt{happy(X)}\\
	$h_3^\texttt{LIKES}$ \> \texttt{LIKES(X,Y)} $\leftarrow$ \texttt{meets(X,Z,Y)}, \texttt{RICH(Z)}
\end{tabbing}  
belonging to $\mathcal{L}^\texttt{LIKES}$ can be considered hypotheses of a definition for \texttt{LIKES}.
\end{example}

The \emph{set $\mathcal{E}$ of examples} in Definition \ref{def:learning-problem} contains assertions of the kind $p(\vec{a_i})$ where $p$ is the target predicate and $\vec{a_i}$ is a tuple of individuals occurring in the ABox $\mathcal{A}$. Note that, when $p$ is a role name, the tuple $\vec{a_i}$ is a pair $<a_i^1, a_i^2>$ of individuals. We assume $\mathcal{B} \cap \mathcal{E} = \emptyset$. However, a possibly incomplete description of each $e_i \in \mathcal{E}$ is in $\mathcal{B}$.
\begin{example}\label{ex:observations2}
With reference to Example \ref{ex:hyp-lang2}, suppose that the following concept assertions:
\begin{tabbing}
  MMMMMM \= MMMMMM \= MM \= \kill
	$e_1^\texttt{LONER}$ \> \texttt{LONER(Mary)}\\
	$e_2^\texttt{LONER}$ \> \texttt{LONER(Joe)}\\
	$e_3^\texttt{LONER}$ \> \texttt{LONER(Paul)}
\end{tabbing}  
are examples for the target predicate \texttt{LONER}.
\end{example}
\begin{example}\label{ex:observations3}
With reference to Example \ref{ex:hyp-lang3}, the following role assertions:
\begin{tabbing}
  MMMMMM \= MMMMMM \= MM \= \kill
	$e_1^\texttt{LIKES}$ \> \texttt{LIKES(Mary,Italy)}\\
	$e_2^\texttt{LIKES}$ \> \texttt{LIKES(Mary,Germany)}\\
	$e_3^\texttt{LIKES}$ \> \texttt{LIKES(Joe,Italy)}
\end{tabbing}  
can be assumed as examples for the target predicate \texttt{LIKES}.
\end{example}

\subsection{The ingredients for an ILP solution}\label{sect:ilp-solution}

In order to solve the learning problem in hand with the ILP methodological approach , the language $\mathcal{L}$ of hypotheses needs to be equipped with (i) a \emph{coverage relation} which defines the mappings from $\mathcal{L}$ to the set $\mathcal{E}$ of examples, and (ii) a \emph{generality order} $\succeq$ such that $(\mathcal{L}, \succeq)$ is a search space. 

The definition of a \emph{coverage relation} depends on the representation choice for examples. The normal ILP setting is the most appropriate to the learning problem in hand and can be extended to the \negDLlog\ framework depicted in Definition \ref{def:learning-problem} as follows.
\begin{definition}\label{def:coverage_int1}
We say that a rule $h \in \mathcal{L}$ \emph{covers} (does not cover, resp.) an example $e_i = p(\vec{a_i}) \in \mathcal{E}$ w.r.t. a background theory $\mathcal{B}$ iff $\mathcal{B} \cup h \models p(\vec{a_i})$ ($\mathcal{B} \cup h \not\models p(\vec{a_i})$, resp.). 
\end{definition} 
Note that the coverage test can be reduced to query answering w.r.t. a \disjnegDLlog\ KB, which in turn can be reformulated as a satisfiability problem of the KB.

\begin{example}\label{ex:coverage-test2}
With reference to Example \ref{ex:hyp-lang2} and \ref{ex:observations2}, the rule $h_1^\texttt{LONER}$ covers the example $e_1^\texttt{LONER}$ because all NM-models for $\mathcal{B}^\prime = \mathcal{B} \cup h_1^\texttt{LONER}$ do satisfy \texttt{famous(Mary)}. It covers also $e_2^\texttt{LONER}$ and $e_3^\texttt{LONER}$ for analogous reasons.
The rule $h_2^\texttt{LONER}$ covers only $e_1^\texttt{LONER}$ and $e_2^\texttt{LONER}$ whereas $h_3^\texttt{LONER}$ covers $e_2^\texttt{LONER}$ and $e_3^\texttt{LONER}$.
\end{example}
\begin{example}\label{ex:coverage-test3}
With reference to Example \ref{ex:hyp-lang3} and \ref{ex:observations3}, the rule $h_1^\texttt{LIKES}$ covers the example $e_1^\texttt{LIKES}$ because all NM-models for $\mathcal{B}^\prime = \mathcal{B} \cup h_1^\texttt{LIKES}$ do satisfy  \texttt{meets(Mary,Z,Italy)}. It covers also $e_2^\texttt{LIKES}$ and $e_3^\texttt{LIKES}$ for analogous reasons. The rule $h_2^\texttt{LIKES}$ covers only $e_1^\texttt{LIKES}$ and $e_2^\texttt{LIKES}$ whereas $h_3^\texttt{LIKES}$ covers only $e_1^\texttt{LIKES}$ and $e_3^\texttt{LIKES}$.
\end{example}

The definition of a \emph{generality order} for hypotheses in $\mathcal{L}$ must consider the peculiarities of the chosen $\mathcal{L}$.
Generalized subsumption, subsequently extended in \cite{Sakama01} to deal with NAF literals, is suitable for the problem in hand and can be adapted to the case of \negDLlog\ rules. In the following we provide a characterization of the resulting generality order, denoted by $\succeq_{\mathcal{K}}^{\neg}$, that relies on the reasoning tasks known for \disjnegDLlog and from which a test procedure can be derived.   
\begin{definition}\label{def:K-subsumption-2}
Let $h_1, h_2 \in \mathcal{L}$ be two \negDLlog\ rules standardized apart, $\mathcal{K}$ a \negDLlog\ KB, and $\sigma$ a Skolem substitution for $h_2$ with respect to $\{h_1\} \cup \mathcal{K}$. We say that $h_1$ is \emph{more general than} $h_2$ w.r.t. $\mathcal{K}$, denoted by $h_1 \succeq_{\mathcal{K}}^{\neg} h_2$, iff there
exists a ground substitution $\theta$ for $h_1$ such that (i)
$head(h_1)\theta=head(h_2)\sigma$ and (ii) $\mathcal{K} \cup body(h_2)\sigma \models
body(h_1)\theta$. We say that $h_1$ is \emph{strictly more general than} $h_2$ w.r.t. $\mathcal{K}$, denoted by $h_1 \succ_{\mathcal{K}}^{\neg} h_2$, iff $h_1 \succeq_{\mathcal{K}}^{\neg} h_2$ and $h_2 \not\succeq_{\mathcal{K}}^{\neg} h_1$. We say that $h_1$ is \emph{equivalent to} $h_2$ w.r.t. $\mathcal{K}$, denoted by $h_1 \equiv_{\mathcal{K}}^{\neg} h_2$, iff $h_1 \succeq_{\mathcal{K}}^{\neg} h_2$ and $h_2 \succeq_{\mathcal{K}}^{\neg} h_1$. 
\end{definition}

\begin{example}\label{ex:generality-test2}
Let us consider the rules reported in Example \ref{ex:hyp-lang2} up to variable renaming:
\begin{tabbing}
  MMMMMM \= MMMMMM \= MM \= \kill
	$h_1^\texttt{LONER}$ \> \texttt{LONER(A)} $\leftarrow$ \texttt{famous(A)}\\
	$h_2^\texttt{LONER}$ \> \texttt{LONER(X)} $\leftarrow$ \texttt{famous(X)},\texttt{UNMARRIED(X)}
\end{tabbing}
In order to check whether $h_1^\texttt{LONER} \succeq_{\mathcal{K}}^{\neg} h_2^\texttt{LONER}$ holds, let $\sigma = \{\texttt{X}/\texttt{a}\}$ a Skolem substitution for $h_2^\texttt{LONER}$ with respect to $\mathcal{K} \cup h_1^\texttt{LONER}$ and $\theta =\{\texttt{A}/\texttt{a}\}$ a ground substitution for $h_1^\texttt{LONER}$. The condition (i) is immediately verified. The condition
\begin{center}
 (ii)  $\mathcal{K} \cup \{\texttt{famous(a)},\texttt{UNMARRIED(a)}\} \models \{\texttt{famous(a)}\}$ 
\end{center}
is a ground query answering problem in \negDLlog. It can be easily proved that all NM-models for $\mathcal{K} \cup \{\texttt{famous(a)},\texttt{UNMARRIED(a)}\}$ satisfy $\texttt{famous(a)}$. Thus, it is the case that $h_1^\texttt{LONER} \succeq_{\mathcal{K}}^{\neg} h_2^\texttt{LONER}$. The viceversa does not hold. Also, $h_1^\texttt{LONER} \succ_{\mathcal{K}}^{\neg} h_3^\texttt{LONER}$ and $h_3^\texttt{LONER}$ is incomparable with $h_2^\texttt{LONER}$.
\end{example}
\begin{example}\label{ex:generality-test3}
With reference to Example \ref{ex:hyp-lang3}, it can be proved that $h_1^\texttt{LIKES} \succ_{\mathcal{K}}^{\neg} h_2^\texttt{LIKES}$ and $h_1^\texttt{LIKES} \succ_{\mathcal{K}}^{\neg} h_3^\texttt{LIKES}$. Conversely, the rules $h_2^\texttt{LIKES}$ and $h_3^\texttt{LIKES}$ are incomparable. Note that
\begin{tabbing}
  MMMMMM \= MMMMMM \= MM \= \kill
	$h_4^\texttt{LIKES}$ \> \texttt{LIKES(X,Y)} $\leftarrow$ \texttt{meets(X,Z,Y)}, \texttt{LOVES(X,Z)}\\
	$h_5^\texttt{LIKES}$ \> \texttt{LIKES(X,Y)} $\leftarrow$ \texttt{meets(X,Z,Y)}, \texttt{WANTS-TO-MARRY(X,Z)}
\end{tabbing}
also belong to $\mathcal{L}^\texttt{LIKES}$. It can be proved that $h_1^\texttt{LIKES} \succ_{\mathcal{K}}^{\neg} h_4^\texttt{LIKES}$, $h_1^\texttt{LIKES} \succ_{\mathcal{K}}^{\neg} h_5^\texttt{LIKES}$, and $h_4^\texttt{LIKES} \succ_{\mathcal{K}}^{\neg} h_5^\texttt{LIKES}$.
\end{example}

Note that the decidability of $\succ_{\mathcal{K}}^{\neg}$ follows from the decidability of \negDLlog.
Also it can be proved that $\succ_{\mathcal{K}}^{\neg}$ is a quasi-order (\ie, it is a reflexive and
transitive relation) for \negDLlog\ rules, therefore the space $(\mathcal{L}, \succ_{\mathcal{K}}^{\neg})$ can be searched by refinement operators like the following one able to traverse the hypothesis space top down. 

\begin{definition}\label{def:ref-op-1}
Let $\mathcal{L}$ be a \negDLlog\ hypothesis language built out of the three finite and disjoint alphabets $P_\mathcal{C}(\mathcal{L})$, $P_\mathcal{R}(\mathcal{L})$, and $P_\textsc{D}^{+}(\mathcal{L}) \cup P_\textsc{D}^{-}(\mathcal{L})$.
We define a \emph{downward refinement operator} $\rho^{\textsc{OR}}$ for
$(\mathcal{L}, \succeq_{\mathcal{K}}^{\neg})$ such that, for each $h \in \mathcal{L}$, the set
$\rho^{\textsc{OR}}(h)$ contains all $h^\prime \in\mathcal{L}$ that can be obtained from $h$ by
applying one of the following refinement rules:
\begin{description}
  \item [$\langle AddDataLit\_B^{+} \rangle$] $body(h^\prime)=body(h) \cup \{r_{m+1}(\vec{Y_{m+1}})\}$ if
		\begin{enumerate}
			\item $r_{m+1} \in P_\textsc{D}^{+}(\mathcal{L})$
			\item $r_{m+1}(\vec{Y_{m+1}})\not\in body(h)$
			\item $var(head(h)) \subseteq var(body(h^\prime))$
		\end{enumerate}
\end{description}
\begin{description}
  \item [$\langle AddOntoLit\_B \rangle$] $body(h^\prime)=body(h) \cup \{s_{k+1}(\vec{Z_{k+1}})\}$ if
  	\begin{enumerate}
			\item $s_{k+1} \in P_\mathcal{C}(\mathcal{L}) \cup P_\mathcal{R}(\mathcal{L})$
			\item it does not exist any $s_{l}(\vec{Z_{l}})\in body(h)$ such that $s_{k+1} \sqsubseteq s_{l}$
			\item $var(head(h)) \subseteq var(body(h^\prime))$
		\end{enumerate}
\end{description}
\begin{description}
  \item [$\langle SpecOntoLit\_B \rangle$] $body(h^\prime)= (body(h) \setminus \{s_{l}(\vec{Z_{l}})\}) \cup s_{l}^\prime(\vec{Z_{l}})$ if
  	\begin{enumerate}
			\item $s_{l}^\prime \in P_\mathcal{C}(\mathcal{L}) \cup P_\mathcal{R}(\mathcal{L})$
			\item $s_{l}^\prime \sqsubseteq s_{l}$
		\end{enumerate}
\end{description}
\begin{description}
  \item [$\langle AddDataLit\_B^{-} \rangle$] $body(h^\prime)=body(h) \cup \{\neg u_{q+1}(\vec{W_{q+1}})\}$ if
		\begin{enumerate}
			\item $u_{q+1} \in P_\textsc{D}^{-}(\mathcal{L})$
			\item $u_{q+1}(\vec{W_{q+1}})\not\in body(h)$
			\item $\vec{W_{q+1}} \subset var(body^{+}(h))$
		\end{enumerate}
\end{description}
\end{definition}

All the rules of $\rho^{\textsc{OR}}$ are correct, \ie\ the $h^\prime$'s obtained by applying any of the
rules of $\rho^{\textsc{OR}}$ to $h \in \mathcal{L}$ are such that $h \succ_{\mathcal{K}}^{\neg} h^\prime$. This can be
proved intuitively by observing that they act only on $body(h)$. Thus condition (i) of
Definition \ref{def:K-subsumption-2} is satisfied. Furthermore, it is straightforward
to notice that the application of any of the
rules of $\rho^{\textsc{OR}}$ to $h$ reduces the number of
models of $h$. In particular, as for $\langle SpecOntoLit\_B \rangle$, this
intuition follows from the semantics of DLs. So condition (ii) also is fulfilled.
\begin{example}\label{ex:ref-op-2}
With reference to Example \ref{ex:hyp-lang2}, applying $\langle AddDataLit\_B^{+} \rangle$ to
\begin{tabbing}
  MMMMMM \= MMMMMM \= MM \= \kill
	$h_0^\texttt{LONER}$ \> \texttt{LONER(X)} $\leftarrow$
\end{tabbing}  
produces $h_1^\texttt{LONER}$ which can be further specialized by means of $\langle AddOntoLit\_B \rangle$ and $\langle AddDataLit\_B^{-} \rangle$. 
Note that no other refinement rule can be applied to $h_1^\texttt{LONER}$ and that $h_2^\texttt{LONER}$ and $h_3^\texttt{LONER}$ are among the refinements of $h_1^\texttt{LONER}$. 
\end{example}
\begin{example}\label{ex:ref-op-3}
With reference to Example \ref{ex:hyp-lang3}, applying $\langle AddDataLit\_B^{+} \rangle$ to
\begin{tabbing}
  MMMMMM \= MMMMMM \= MM \= \kill
	$h_0^\texttt{LIKES}$ \> \texttt{LIKES(X,Y)} $\leftarrow$
\end{tabbing}  
produces $h_1^\texttt{LIKES}$ which can be further specialized into $h_2^\texttt{LIKES}$, $h_3^\texttt{LIKES}$, $h_4^\texttt{LIKES}$ and $h_5^\texttt{LIKES}$ by means of $\langle AddDataLit\_B \rangle$ and $\langle AddOntoLit\_B \rangle$. 
Note that no other refinement rule can be applied to $h_1^\texttt{LIKES}$ and that $h_5^\texttt{LIKES}$ can be also obtained as refinement from $h_4^\texttt{LIKES}$ via $\langle SpecOntoLit\_B \rangle$. 
\end{example}

\subsection{An ILP algorithm}\label{sect:foil-like-algo}

\begin{figure}[t]
  \centering
\begin{tabbing}
    MMM \= MMM \= MMM \= MMM \= MMM \kill
    \textbf{function} \ORFoil($\mathcal{B}$, $p$, $\mathcal{E}^{+}$, $\mathcal{E}^{-}$, $\mathcal{L}$): $\mathcal{H}$\\
    1.  $\mathcal{H} := \emptyset$\\
    2.  \textbf{while} $\mathcal{E}^{+} \neq \emptyset$ \textbf{do}\\
		3.	\> $h := \{ p(\vec{X}) \leftarrow \}$;\\
		4.	\> $\mathcal{E}^{-}_{h} := \mathcal{E}^{-}$\\
    5.  \> \textbf{while} $\mathcal{E}^{-}_{h} \neq \emptyset$ \textbf{do}\\
    6.  \> \> $\mathcal{Q} := \{ h^\prime \in \mathcal{L}| h^\prime \in \rho^{\textsc{OR}}(h)\}$;\\
		7.	\> \> $h :=$ \ORFoil-\textsc{ChooseBest}$(\mathcal{Q})$;\\
    8.	\> \> $\mathcal{E}^{-}_{h} := \{ e \in \mathcal{E}^{-} | \mathcal{B} \cup h \models e\}$;\\
    9.  \> \textbf{endwhile}\\
    10.	\> $\mathcal{H} := \mathcal{H} \cup \{ h \}$;\\
    11. \> $\mathcal{E}^{+}_{h} := \{ e \in \mathcal{E}^{+} | \mathcal{B} \cup h \models e\}$;\\
    12. \> $\mathcal{E}^{+} := \mathcal{E}^{+} \setminus \mathcal{E}^{+}_{h}$\\	
    13. \textbf{endwhile}\\
    14. \textbf{return} $\mathcal{H}$
\end{tabbing}
  \caption{\ORFoil: A \Foil-like algorithm for learning onto-relational rules}\label{fig:foil-like-algo}
\end{figure}

The ingredients identified in the previous section are the starting point for the definition of ILP algorithms. Figure \ref{fig:foil-like-algo} reports the main procedure of a \Foil-like algorithm, named \ORFoil, for learning onto-relational rules. In \ORFoil, analogously to \Foil\footnote{\Foil\ is a popular ILP algorithm for learning sets of rules to be used as a classifier \cite{Quinlan90}.}, the outer loop (steps 2-12) corresponds to a variant of the sequential covering algorithm, \ie, it learns new rules one at a time, removing the positive examples covered by the latest rule before attempting to learn the next rule (steps 11-12). The hypothesis space search performed by \ORFoil\ is best understood by viewing it hierarchically. Each iteration through the outer loop (steps 2-13) adds a new rule to its disjunctive hypothesis $\mathcal{H}$. The effect of each new rule is to generate the current disjunctive hypothesis (\ie, to increase the number of instances it classifies as positive), by adding a new disjunct. Viewed at this level, the search is a bottom-up search through the space of hypotheses, beginning with the most specific empty disjunction (step 1) and terminating when the hypothesis is sufficiently general to cover all positive training examples (step 13). The inner loop (steps 5-9) performs a more fine-grained search to determine the exact definition of each new rule. This loop searches a second hypothesis space, consisting of conjunctions of literals, to find a conjunction that will form the body of the new rule. Within this space, it conducts a top-down, hill-climbing search, beginning with the most general preconditions possible (step 3), then refining the rule (step 6) until it avoids all negative examples. To select the most promising specialization from the candidates generated at each iteration, \ORFoil-\textsc{ChooseBest} (called at step 7) considers the performance of each candidate over $\mathcal{E}$ and chooses the one which maximizes the \emph{information gain}.
This measure is computed according to the following formula
\begin{equation}\label{eq:gain}
	\textsc{Gain}(h^{\prime}, h) = p \ast (log_2(cf(h^{\prime})) - log_2(cf(h))) \ , 
\end{equation}
where $p$ is the number of distinct variable bindings with which positive examples covered by the
rule $h$ are still covered by $h^{\prime}$ and $cf()$ is the confidence degree.
Thus, the gain is positive iff $h^{\prime}$
is more informative in the sense of Shannon's information
theory (\ie\ iff the confidence degree increases). If
there are some literals to add which increase the confidence
degree, the information gain tends to favor the literals that offer the best
compromise between the confidence degree and the number
of examples covered.

One may think to use the confidence degree defined for \DLFoil\ (see Chapter \ref{OL-book-chapter-concept-learning} for more details) which takes OWA into account. Indeed, many individuals may be available which can not be classified as instances of the target concept nor of its negation. This requires a different setting able to deal with unlabeled individuals.

\begin{example}\label{ex:foil-like-algo-2}
With reference to Example \ref{ex:ref-op-2} and Example \ref{ex:coverage-test2}, we suppose that 
\begin{tabbing}
  MM \= MMMMMM \= MM \= \kill
\> $\mathcal{E}^{+} = \{e_1^\texttt{LONER}, e_2^\texttt{LONER}\}$\\
\> $\mathcal{E}^{-} = \{e_3^\texttt{LONER}\}$
\end{tabbing}
The outer loop of \ORFoil\ starts from $h_0^\texttt{LONER}$ which is further refined through the iterations of the inner loop, more precisely it is first specialized into $h_1^\texttt{LONER}$ which in turn, since it covers negative examples, is then specialized into $h_2^\texttt{LONER}$ and $h_3^\texttt{LONER}$ out of which the rule $h_3^\texttt{LONER}$ is added to $\mathcal{H}^\texttt{LONER}$ the hypothesis because it does not cover negative examples. At this point the algorithm stops because $\mathcal{H}^\texttt{LONER}$ covers both positive examples.
\end{example}
\begin{example}\label{ex:ex:foil-like-algo-3}
Following Example \ref{ex:ref-op-3} and Example \ref{ex:coverage-test3}, we assume that $\mathcal{E}^{+} = \{e_1^\texttt{LIKES}, e_3^\texttt{LIKES}\}$ and $\mathcal{E}^{-} = \{e_2^\texttt{LIKES}\}$. At the end of the first iteration, $h_3^\texttt{LIKES}$ is included into $\mathcal{H}^\texttt{LIKES}$ since it does not cover negative examples but only one positive example.
\end{example}

\section{Final Remarks and Directions of Research}\label{sect:concl}

Building rules within ontologies poses several challenges not only to KR researchers investigating suitable hybrid DL-CL formalisms but also to the ML community which has been historically interested in application areas where the knowledge acquisition bottleneck is particularly severe.
In particular, ORL may open up new opportunities for KE because it will make systems available to support the knowledge engineer in her most demanding task, \ie\ defining rules that extend or complement an ontology. Thus, ORL may produce time and cost savings in KE. 
In this chapter, we have revised the ML literature addressing the problem of learning onto-relational rules. Very few ILP works have been found that propose a solution to this problem \cite{RouveirolV2000,Lisi08,LisiE08-ilp}. They adopt \Carin-$\mathcal{ALN}$, \ALlog\ and \SHIQlog\ as KR framework, respectively. Note that matching Table \ref{tab:ilp-comp} against Table \ref{tab:kr-comp} one may figure out what is the state-of-the-art and what are the directions of research on onto-relational rules from the ML viewpoint. Also he/she can get suggestions on what is the most appropriate among these ILP frameworks to be implemented for a certain intended application. The specific solution illustrated in Section \ref{sect:learning-fwk} takes advantage from an augmented expressive power thanks to the chosen \disjnegDLlog\ instantiation \cite{DBLP:journals/tplp/Lisi10}. It supports the evolution of ontologies with the creation of a concept/role, change operations which both boil down to the addition of new rules to the input KB. 

From the comparative analysis of the ILP frameworks reviewed in Section \ref{sect:ilp4sw-today}, a common feature emerges: All proposals resort to Buntine's generalized subsumption and extend it in a non-trivial way. This choice is due to the fact that, among the semantic generality orders in ILP, generalized subsumption applies only to definite clauses, therefore suits well the hypothesis language in all three frameworks. Following these guidelines, new ILP frameworks can be designed to deal with more or differently expressive hybrid DL-CL languages according to the DL chosen (\eg, learning \Carin-$\mathcal{ALCNR}$ rules), or the clausal language chosen (\eg, learning recursive \Carin\ rules), or the integration scheme (\eg, learning \Carin\ rules with \DL-literals in the head). An important requirement will be the definition of a \emph{semantic} generality relation for hypotheses to take into account the background knowledge. Of course, generalized subsumption may turn out to be not suitable for all cases, \eg\ for the case of learning \disjDLlog\ rules \cite{DBLP:journals/tplp/Lisi10}. 
Also it would be interesting to investigate how the nature of rules (\ie, the intended context of usage) may impact the learning process as for the scope of induction and other variables in the learning problem statement. For example, the problem of learning \ALlog\ rules for classification purposes differ greatly from the apparently similar learning problem faced in \cite{LisiM04}. Finally, it is worthy to consider hybrid KR formalisms with loose and full integration scheme. 

Besides theoretical issues, most future work will have to be devoted to implementation and application. When moving to practice, issues like efficiency and scalability become of paramount importance. These concerns may drive the attention of ILP research towards less expressive hybrid KR frameworks in order to gain in tractability, \eg\ instantiations of \disjnegDLlog\ with DL-Lite \cite{CalvaneseLRV04}. Applications can come out of some of the many use cases for Semantic Web rules specified by the RIF W3C Working Group. 





%

\end{document}